\documentclass[10pt,twocolumn,letterpaper]{article}

\usepackage[pagenumbers]{cvpr} 

\usepackage{graphicx}
\usepackage{amsmath}
\usepackage{amssymb}
\usepackage{booktabs}
\usepackage[T1]{fontenc}

\newcommand{\modelName}{Manipulation via Visual Object Location Estimation\xspace}
\newcommand{\modelShort}{m-VOLE\xspace}

\providecommand{\taskName}{Object Displacement}
\providecommand{\taskshort}{ObjDis}
\newcommand{\thor}{\textsc{AI2-Thor}\xspace}

\newcommand{\armpnav}{\textsc{ArmPointNav}\xspace}

\usepackage{multirow}
\usepackage{enumerate}
\usepackage{color}
\usepackage{xspace}
\usepackage{booktabs}
\usepackage{makecell}

\usepackage{enumitem}

\usepackage[pagebackref,breaklinks,colorlinks]{hyperref}

\usepackage[capitalize]{cleveref}
\crefname{section}{Sec.}{Secs.}
\Crefname{section}{Section}{Sections}
\Crefname{table}{Table}{Tables}
\crefname{table}{Tab.}{Tabs.}

\def\confName{CVPR}
\def\confYear{2022}

\begin{document}

\title{Object Manipulation via Visual Target Localization}

\author{Kiana Ehsani$^{1}$ \xspace\xspace
Ali Farhadi$^{2}$\xspace\xspace
Aniruddha Kembhavi$^{1,2}$\xspace\xspace
Roozbeh Mottaghi$^{1,2}$\\
$^1$ Allen Institute for AI\xspace \xspace $^2$ University of Washington
}

\twocolumn[{
\renewcommand\twocolumn[1][]{#1}
\maketitle
\centering
\includegraphics[width=\linewidth]{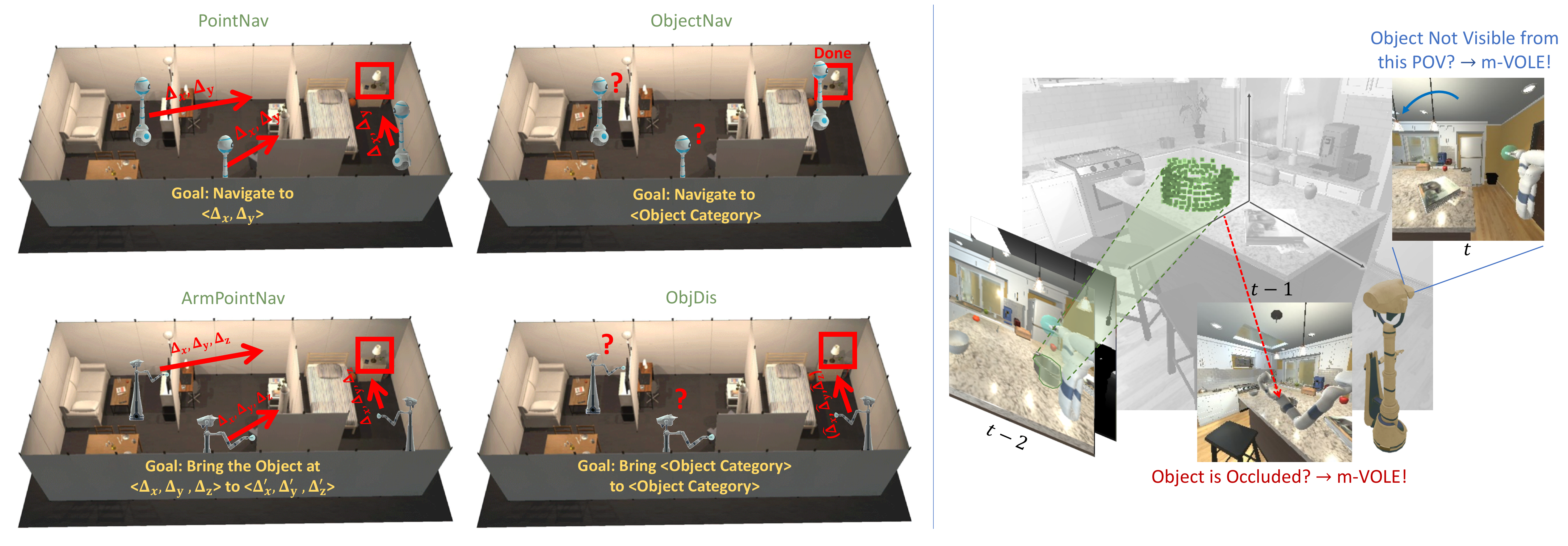}
\captionof{figure}{We propose to solve a manipulation task, \taskName\ (\taskshort), where the goal is to bring a source object towards a destination object (e.g., bring a bowl to sink). In previous popular Embodied AI tasks shown in the left panel, either the goal is defined by the relative 3D coordinates  (PointNav~\cite{savva2019habitat} and ArmPointNav~\cite{manipulathor}) or there is no manipulation involved (ObjectNav~\cite{batra2020objectnav}). In contrast, we estimate the goal location from visual observations and manipulate objects across the scenes. The right panel shows an example that the agent robustly estimates the relative object location despite the occlusion by the arm and being out of view. }
\label{fig:teaser}
\vspace*{0.5cm}
}] 

\maketitle
\thispagestyle{empty}


\begin{abstract}
Object manipulation is a critical skill required for Embodied AI agents interacting with the world around them. Training agents to manipulate objects, poses many challenges. These include occlusion of the target object by the agent's arm, noisy object detection and localization, and the target frequently going out of view as the agent moves around in the scene. We propose \modelName (\modelShort), an approach that explores the environment in search for target objects, computes their 3D coordinates once they are located, and then continues to estimate their 3D locations even when the objects are not visible, thus robustly aiding the task of manipulating these objects throughout the episode. Our evaluations show a massive 3\emph{x} improvement in success rate over a model that has access to the same sensory suite but is trained without the object location estimator, and our analysis shows that our agent is robust to noise in depth perception and agent localization. Importantly, our proposed approach relaxes several assumptions about idealized localization and perception that are commonly employed by recent works in embodied AI -- an important step towards training agents for object manipulation in the real world.

\end{abstract}

\section{Introduction}
\label{sec:intro}
In recent years the computer vision community has made steady progress on a variety of Embodied AI tasks including navigation \cite{Anderson2018OnEO,batra2020objectnav,zhu2017target,chaplot2020object}, object manipulation \cite{manipulathor,xia2020relmogen,szot2021habitat,Weihs2021VisualRR} and language-based tasks \cite{Shridhar2020ALFREDAB,Anderson2018VisionandLanguageNI,gordon2018iqa,das2018embodied} within interactive worlds in simulation \cite{kolve2017ai2,savva2019habitat,Gan2020ThreeDWorldAP,Xia2020InteractiveGB}. Performances of state-of-the-art systems on these tasks vary greatly depending on the complexity of the task, the assumptions made about the agent and environment, and the sensors employed by the agent. While robust and reliable methods have emerged for some of these tasks, developing generalizable and scalable solutions remains a topic of research. A challenging problem in this domain is visual object manipulation, a critical skill that enables the agents to interact with objects and change the world around them. Avoiding the collision of the arm with other objects in the scene, inferring the state of the scene using noisy visual observations, and planning an efficient path for the agent and its arm towards objects are a few of many interesting challenges in this domain.

One of the main obstacles in training capable and generalizable agents for visual object manipulation is the sparsity of training signals. Early works in visual navigation in the Embodied AI (EAI) research community (e.g., \cite{zhu2017target}) suffered from this training signal problem. To alleviate this issue, the EAI community provided the agent with a powerful suite of sensors. The first incarnations of the Point Goal Navigation (PointNav) task (navigating to a specified X-Y coordinate within an environment) relied on perfect sensory information that included GPS localization for the target along with compass and GPS sensors for the agent \cite{savva2019habitat}. As a result, the agent was able to access accurate relative coordinates of its goal at every time step, which acted as a dense supervisory signal during training and inference. Under these assumptions, models with minimal inductive biases achieve near-perfect accuracy \cite{Wijmans2020DDPPOLN} in unseen test environments given enough training. Mirroring this success story, researchers have also been able to train effective agents for visual object manipulation \cite{manipulathor}. 

While these are promising steps towards building embodied agents, the strong sensory assumptions employed by these models limit their applicability in the real world. Their task definition requires specifying the target via 3D coordinates, which can be extremely hard to determine, if not impossible, in indoor environments. Moreover, they rely on a perfect compass and GPS sensory information, which enables the agent to localize itself with respect to the target. In this paper, we take a step closer to a more realistic task definition for object manipulation by specifying the goal via a representative image of a particular object category (instead of their 3D coordinates) and propose a method to localize the target object without relying on any perfect localization sensory information. 

We introduce \taskName\ (\taskshort), the task of bringing an object to a target location (e.g., \textit{bring a bowl to the sink}). This involves searching for an object, navigating to it, picking it up, and placing it at the desired target location. Figure~\ref{fig:teaser}(left) contrasts \taskshort\ with other popular navigation and manipulation tasks. We propose \modelName (\modelShort), a model for \taskshort\ that continually estimates the 3D relative location of the target to the agent via visual observations and learns a policy to perform the desired manipulation task. \modelShort predicts a segmentation mask for the objects of interest and leverages the depth sensor to estimate the relative 3D coordinates for these objects. However, target segmentations are not always available to the agent due to several reasons including: objects may be out of view due to the agent being in a different location, objects may be occluded by the arm of the agent, and masks may be unavailable due to imperfect object segmentation models; and these noisy observations of the target can lead to the failure of the agent. To alleviate these issues, \modelShort aggregates estimates for the objects' 3D coordinates over time, leverages previous estimates when the object mask is unavailable, and can seamlessly re-localize the object in its coordinate frame when the agent observes it again -- rendering the model robust to noise in movement and perception. Figure~\ref{fig:teaser}(right) shows a schematic of the agent's object localization in the presence of occlusion.

We conduct our experiments using the ManipulaTHOR~\cite{manipulathor} framework which provides visually rich, physics-enabled environments with a variety of objects. We show that:
\vspace{-0.7em}

\begin{enumerate}[label=(\alph*)]
\itemsep-0.1em
\item Our model achieves 3x success rate compared to a model that is trained without target localization but with an identical set of sensory suites.
\item Our model is robust against noise in perception and agent movements compared to the baselines.
\item Our method allows for zero-shot manipulation of novel objects in unseen environments.
\end{enumerate}

\section{Related Works}
\label{sec:related}
\noindent \textbf{Robotic manipulation.} Manipulation and rearrangement of objects is a long-lasting problem in robotics \cite{lozano1989task,hwang1992gross,kaelbling2013integrated}. Various approaches assume the full visibility of the environment and operate based on the assumption that a perception module provides the perfect observation of the environment \cite{huang2019large,king2016rearrangement,krontiris2015dealing,garrett2018ffrob}. In contrast to these approaches, we focus on the visual perception problem and solve the task when the agent has partial and noisy observations. Several approaches have been proposed (e.g., \cite{Danielczuk2021ObjectRU,yen2020learning,mahler2017learning,zeng2018robotic}) that address visual perception as well. However, the mentioned works focus on the tabletop rearrangement of objects. In contrast, we consider mobile manipulation of objects, which is a more general problem. Mobile manipulation has been explored in the literature as well. However, they typically use a single environment to develop and evaluate the models \cite{stilman2007manipulation,chang2010planning,Nieuwenhuisen2013MobileBP,Nedunuri2014SMTbasedSO}. One of the important problems we address in this paper is the generalization to unseen environments and configurations. \\
\noindent \textbf{Manipulation in virtual environments.} Recently, various works have addressed the problem of object manipulation and rearrangement in virtual environments. These works typically abstract away grasping and, unlike the works mentioned above, mostly focus on visual perception for manipulation, learning-based planning, and generalization to unseen environments and objects. Robosuite \cite{zhu2020robosuite}, RLBench \cite{james2020rlbench} and Meta-world \cite{yu2019meta} provide a set of benchmarks for tabletop manipulation. ManiSkill benchmark \cite{mu2021maniskill} built upon the Sapien framework \cite{xiang2020sapien} is designed to benchmark manipulation skills over diverse objects using a static robotic arm. In contrast to these works, we focus on object displacement (i.e., navigation and manipulation) in unseen scenes (e.g., a kitchen). \cite{xia2020relmogen}  and \cite{szot2021habitat} propose a set of tasks in the iGibson \cite{Shen2020iGibsonAS} and Habitat 2.0 \cite{szot2021habitat} frameworks. However, the same environment and objects are used for train and test. In contrast, our focus is on generalization to unseen environments. ManipulaTHOR~\cite{manipulathor} is an object manipulation framework that employs a robotic arm and introduces a task and benchmark for mobile manipulation. Their task highly depends on various accurate sensory information, which renders the task unrealistic. In contrast to this work, we relax most supervisory signals to better mimic real-world scenarios. \\ 
\noindent \textbf{Relaxing supervisory sensors.} There have been some attempts to relax the assumptions about perfect sensory information using visual odometry for the navigation task \cite{Zhao2021TheSE,Datta2020IntegratingEL}. While these are effective approaches for PointGoal navigation, the same approach does not apply to manipulation as they still require the goal's location in the agent's initial coordinate frame. In contrast, we have no access to the target or goal location at any time step.\\ 
\noindent \textbf{Embodied interactive tasks.}  Navigation is a crucial skill for embodied agents. Various recent techniques have addressed the navigation problem \cite{zhu2017target,yang2018visual,Savinov2018SemiparametricTM,wortsman2019learning,Wijmans2020DDPPOLN,Anderson2018OnEO,gupta2017cognitive,batra2020objectnav,ramakrishnan2020occupancy,chaplot2020object}. These tasks mostly assume the environments are static, and objects do not move. We consider joint manipulation and navigation, which poses different challenges. \cite{Zeng2021PushingIO,Xia2020InteractiveGB} assume objects can move during navigation, but the set of manipulation actions is restricted to push actions. \cite{Anderson2018VisionandLanguageNI,Shridhar2020ALFREDAB,Misra:18goalprediction} propose instruction following to navigate and manipulate objects. They are designed either for static environments \cite{Anderson2018VisionandLanguageNI} or abstract object manipulation \cite{Shridhar2020ALFREDAB,Misra:18goalprediction} (e.g., objects snap to the agent). Works such as \cite{Mirowski2017LearningTN,Marza2021TeachingAH} use auxiliary tasks to overcome the issues related to sparse training signals for navigation. We focus on manipulation, which deals with a different set of challenges. \cite{Weihs2021VisualRR,Batra2020RearrangementAC} propose room rearrangement tasks using mobile robots. Similar to our paper, it involves navigation and displacement of objects. However, their object manipulation is unrealistic in that they assume a magic pointer abstraction, i.e., no arm movement is involved. In contrast, we manipulate objects using an arm, which introduces unique challenges such as planning the motion of the physical arm and avoiding collisions of the arm and object in hand with the rest of the environment.

\section{\taskName}
\label{sec:task}

We introduce the task of \taskName\ (\taskshort), which requires an agent to navigate and manipulate objects in visually rich, cluttered environments. Given two object references, a source object $O_{S}$ and a destination object $O_{D}$, the goal is to locate and pick up $O_{S}$ and move it to the proximity of $O_{D}$ (e.g., \emph{bring an egg to a pot}). The objects of interest are specified to the agent via query images of an instance of each of the categories. Referring to objects via query images as opposed to object names enables us to task the agent with manipulating objects that it has not been trained for. Note that the query images \textit{do not match} the appearance of objects within the scene but are canonical object images (we obtain the images from simulation, but they can also be obtained via other sources such as an image search engine). This enables the user to easily specify the task without the knowledge of object instances in the environment.

The \taskshort\ task consists of multiple implicit stages. To be successful, an agent must (1) explore its environment until it finds the source object, (2) navigate to it, (3) move its arm to the object so that its gripper may pick it up, (4) locate the destination object within the environment (5) navigate to this object and (6) place the object within its gripper in the proximity of the destination object. In visually rich and cluttered environments such as the ones present in \thor~\cite{kolve2017ai2}, this poses several challenges to the agent. \emph{Firstly}, the agent must learn to move its body (navigate) as well as its arm (manipulate) effectively towards objects of interest to complete the desired task. \emph{Secondly}, the agent must avoid collisions with other objects in the scene, which may occur with its body, its arm, or the source object once it has been picked up. \emph{Thirdly}, the agent must be able to plan its actions over long horizons as it involves multiple objects, exploration, and manipulation. \emph{Finally}, the agent must overcome noisy perception caused by frequent obstruction of its view due to its occluding arm, noisy depth sensors, and imperfect visual processing such as object detection.

We situate our experiments in \thor~\cite{kolve2017ai2}, a simulated set of environments built in Unity with a powerful physics engine, and adopt the agent provided by the ManipulaTHOR~\cite{manipulathor} framework. This agent has a rigid body with a 6DOF arm. The agent is capable of moving its body and arm in the environment while satisfying the kinematic constraints of the arm. Grasping is abstracted to a magnetic gripper (i.e., the object can be grasped if the gripper touches the object). While object grasping is a rich and challenging problem, using a magnetic gripper enables us to focus our efforts on other challenges, including exploration, navigation, arm manipulation, and long-horizon planning.

The action space for the agent consists of 11 actions. There are three agent movement actions (\textit{MoveAhead}, \textit{RotateRight}, and \textit{RotateLeft}), two arm base movements (\textit{MoveUp}, \textit{MoveDown}), which move the base of the arm up and down with respect to the body of the agent, and six arm movements that move the gripper in x, y, z directions in agent's coordinate frame. Similar to \cite{manipulathor}, the agent movement actions move/rotate the agent's body by 0.2m/45 degrees, and the arm movements move the base of the arm or the gripper by 5cm. The framework uses inverse kinematic to calculate the final position of each joint.

\section{\modelName}
\label{sec:model}
One of the main challenges of the \taskshort\ task is finding and localizing the object of interest in the scenarios that the object is out of the field of view, is not detected, or is occluded by the arm. We propose a model to find the target, estimate its location, keep track of the location over time, and plan a path to accomplish the task. 

Agents tackling the task of \taskshort\ accept as input ego-centric RGB and Depth observations along with query images for the source and destination objects. They must choose one of 11 actions at each step and follow a long sequence of steps to succeed. In the absence of perfect perception and localization, this task proves quite hard to learn, and models fail to generalize to new environments (as shown in Sec.~\ref{sec:experiment}).

\medskip

\subsection{Estimating Relative 3D Object Location}
\label{sec:3dloc}
We first explain how the relative location of the object of interest is estimated. This is not possible for an unobserved object because its location remains unknown. However, once the object comes into view (while exploring the environment using the policy described later in Sec.~\ref{sec:policy}), the agent can start this estimation. Note that this is in contrast to the \armpnav~\cite{manipulathor} task where the groundtruth location of the target is provided at all time steps.

At time step $t$, if the desired objects are in view, one can generate their segmentation masks (we explain in Sec.~\ref{sec:condseg} how we obtain these masks). We refer to these segmentations as $M_S$ and $M_D$ for the source and destination objects, respectively. Given a segmentation mask and the depth observation, the object's center in 3D is estimated by backprojecting the depth map within the segmentation mask into the 3D local coordinate of the agent. Formally, $\hat{d}_O = \pi(D[M_O], K)$, where $\hat{d}_O$ is the estimated center of the object $O$ in the agent's coordinate frame, $M_O$ is the segmentation mask of the object in the current observation, $D[M_O]$ is the observed depth frame masked by the object's segmentation (i.e., depth values for the visible regions of the object), $K$ is the intrinsic matrix of the camera, and $\pi$ is a function that projects pixels into the 3D space based on the depth values and camera intrinsics.

There are various sources of noise in object instance segmentation that make it challenging to have a reliable estimate of the 3D object location. First, segmentation models are far from perfect. They often have false positives and false negatives. Policies that rely on these masks or a quantity derived from these masks would not be able to produce a reliable sequence of actions. Second, the arm may obstruct the view of the agent as it moves in the scene -- this causes the segmentation mask to periodically disappear. Third, the object might go outside the agent's camera frame after it is observed once and as a result not visible. To overcome these issues, we propose to aggregate this information over time. 

To sequentially aggregate the relative coordinates of the object, a weighted average of the previous and the current estimated 3D distance is used. At each time step (after observing the target once), \modelShort calculates the distance $\hat{d}_O$ in the agent's coordinate frame at time $t$. However, as the agent takes action, its coordinate frame keeps shifting. At each time step that the target object is visible, the agent re-localizes the goal in its current observation frame and can readjust the coordinates. To accurately convert all the past estimates to the current agent frame, we must keep track of the agent's relative location $L^r_{t}$ with respect to its starting location. Since we aim to reduce the reliance on perfect sensing, we evaluate the effect of noise in the location estimation (see Sec.~\ref{sec:experiment}). 

\begin{figure}[tp]
    \centering
    \vspace{-0.2cm}
    \includegraphics[width=20pc]{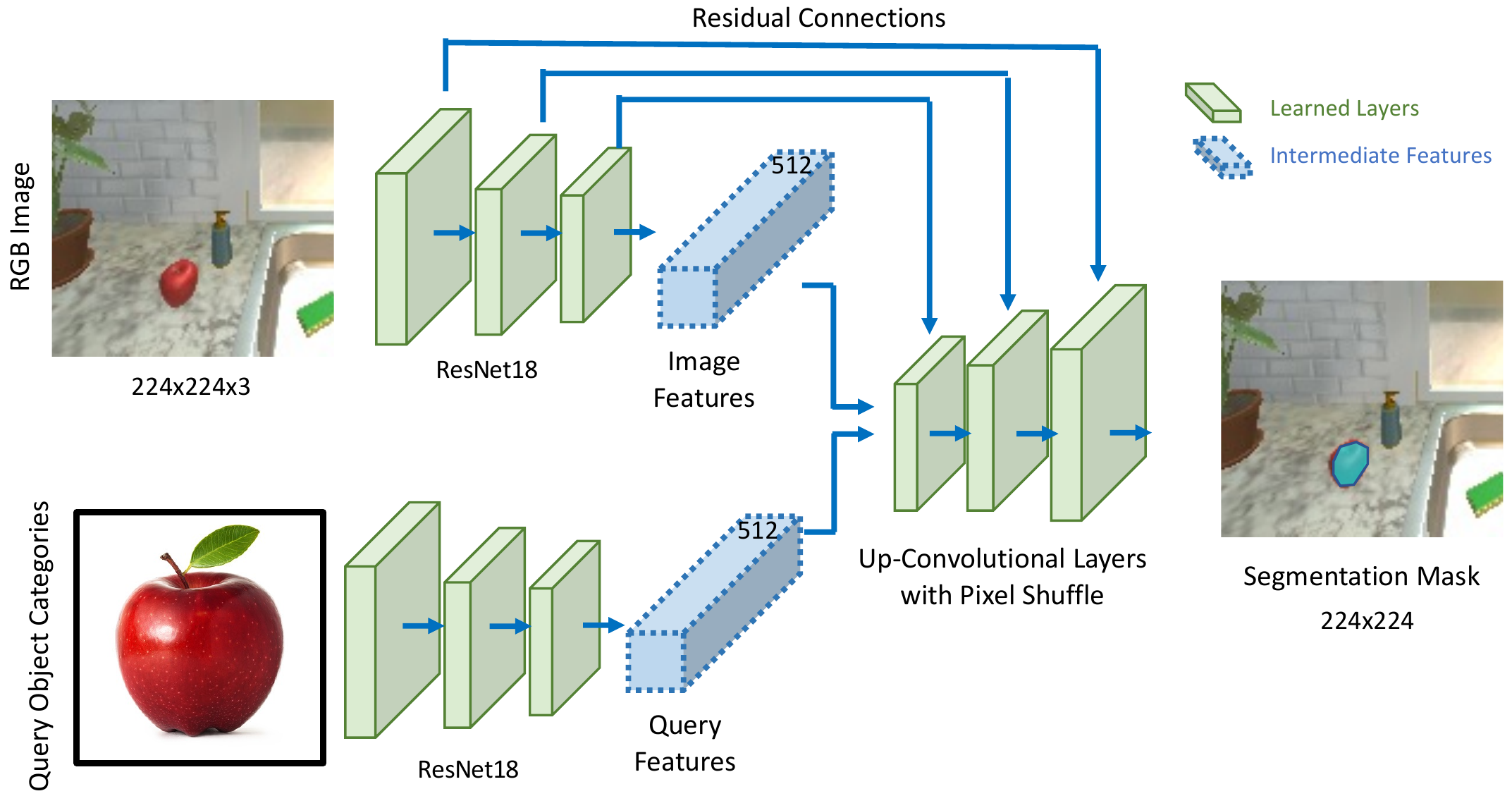}
    \vspace{-0.2cm}
    \caption{\textbf{Conditional Segmentation Architecture}. We use a conditional segmentation model to estimate the segmentation mask. The network receives an RGB image and a query image (representing an object category) as input and outputs the segmentation mask for an instance of that category.}
    \vspace{-0.3cm}
    \label{fig:detection}
\end{figure}

\subsection{Conditional Segmentation}
\label{sec:condseg}
Estimating the 3D location of an observed object requires the agent to estimate its segmentation mask. We refer to this segmentation as `conditional' since the goal is to obtain a mask for an object instance from the object category shown in the conditioning query image. Formulating the problem as a conditional segmentation problem as opposed to the traditional instance segmentation enables the network to focus on generating the mask for the target object\footnote{We use \emph{target object} to refer either the source or destination object.}, and the task reduces to a simpler task of category matching. 

We train an auxiliary segmentation network in an offline setting, independent of the policy network. This allows us to employ a fully supervised setting resulting in efficient policy learning. This network accepts as input two images, the ego-centric RGB observation and a query object, and outputs a mask for all objects from the desired category. This reduces to a binary classification problem at each pixel in the image. The two input images are encoded via ResNet18~\cite{He2016DeepRL} models. The features are then concatenated and passed through five upconvolutional layers. The upconvolutional layers are implemented following PixelShuffle~\cite{Shi2016RealTimeSI}. The final output is a segmentation mask of the same size as the original image (details in Figure~\ref{fig:detection}). We use the Cross-Entropy loss to train this model.

We train and evaluate this model using an offline dataset collected from the train scenes in \thor~\cite{kolve2017ai2}, where we randomize object locations, textures, and scene lighting. We initialize the ResNet18 models with ImageNet pre-training and finetune them on our offline dataset. We compare this method with using a state-of-the-art instance segmentation model trained on the same data in Section~\ref{sec:detection_ablation}.

\subsection{Policy Network}
\label{sec:policy}
\begin{figure*}[h!]
    \centering
    \vspace{-0.5cm}
    \includegraphics[width=30pc]{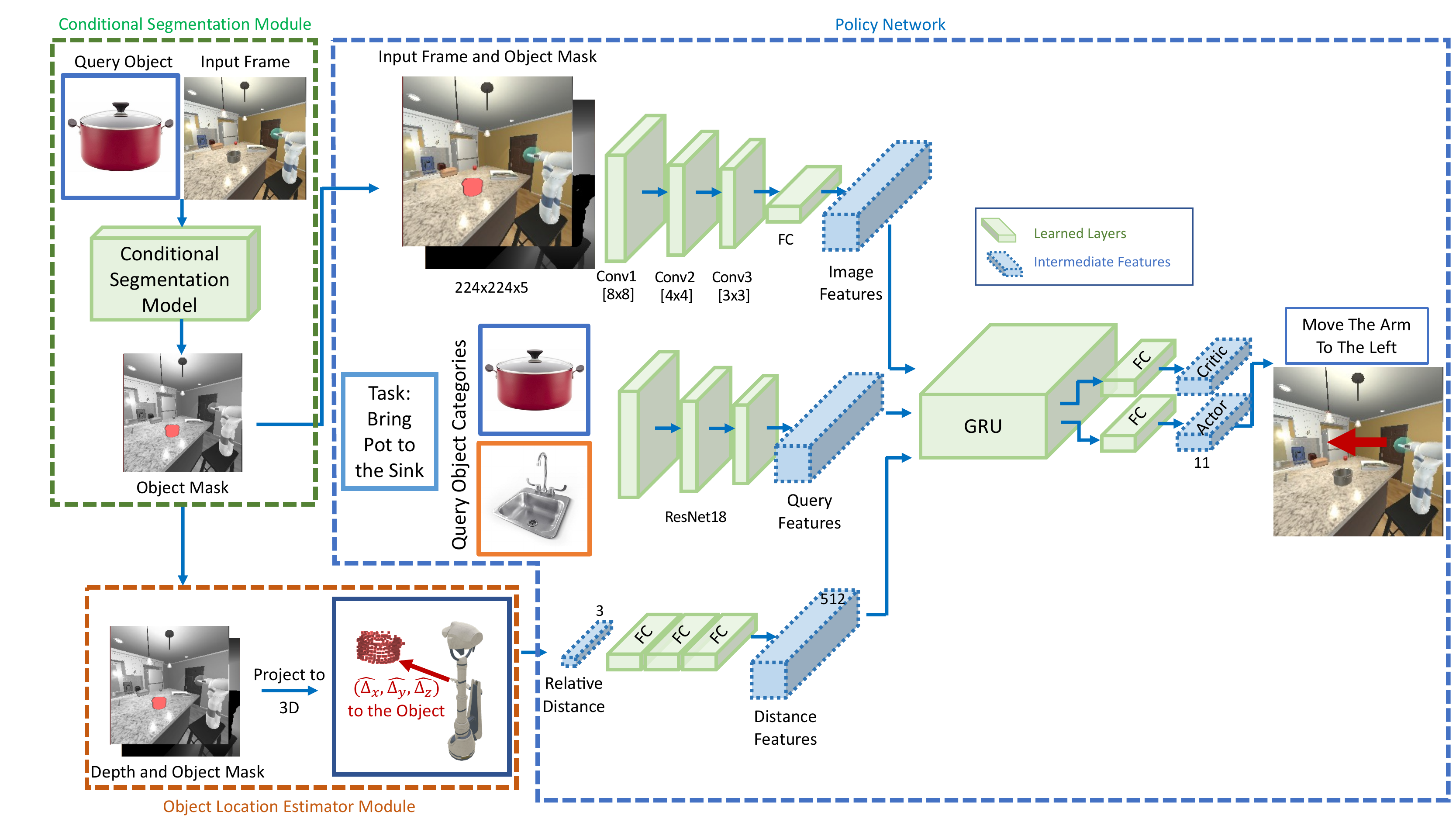}
    \caption{\textbf{Model architecture}. Our model, \modelShort, uses the RGBD observation and a canonical query image of the object to 1) predict the \textit{target object}'s mask (Conditional Segmentation Module), 2) estimate the relative distance of the \textit{target object} in agent's coordinate frame (Object Location Estimator Module), and 3) predict the next action (Policy Network). Note that we use \textit{target object} to refer to either the source or destination object.
}
    \label{fig:policy}
    \vspace{-0.3cm}
\end{figure*}

Figure~\ref{fig:policy} shows a schematic of our proposed model \modelShort. \modelShort\ receives the RGBD observation $I_t$ at the current timestep $t$, and the query images $O_{S}$ and $O_{D}$. The conditional segmentation module estimates the segmentation masks for the target objects $M_S$ and $M_D$ and the visual object location estimator calculates the relative object coordinates $\hat{d}_o$. The policy network then uses $I_t, O_{S}, O_{D}, M_S, M_D, \hat{d}_o$ to sample actions. 

The RGBD observation and segmentation masks of the target objects in the current observation $M_S, M_D$ are concatenated depth-wise and embedded using three convolutional layers. This visual encoding is combined with ResNet18~\cite{He2016DeepRL} (pre-trained on ImageNet~\cite{Russakovsky2015ImageNetLS}) features of the query images of target classes $O_S, O_D$ and the embedding of object's location $\hat{d}_o$. The resulting feature vector is provided to an LSTM, and a final linear layer generates a distribution over the possible actions. We use DD-PPO~\cite{Wijmans2020DDPPOLN} to train the model.  Since our method does not have access to the object's location, and the object is not necessarily initially in sight (only in $13.9$\% of episodes the object is initially visible to the agent), our agent is required to explore the environment until the object is found. We use rewards shown in Table~\ref{tab:reward} to encourage exploration and more efficient manipulation.
In Appendix~\ref{sec:reward_ablation}, we ablate the effects of these rewards on the agent's performance. 

\begin{table}[tp]
\footnotesize
\setlength{\tabcolsep}{3pt}
	\centering
\hfill
	\begin{tabular}{lc|lc|lc}
    \multicolumn{2}{c}{Standard RL Reward}& \multicolumn{2}{c}{Motivating Exploration}&\multicolumn{2}{c}{Efficient Manipulation} \tabularnewline
\toprule 
Failed Action & -0.03 &Object Observed & +1& Pick Up $O_S$ & +5 \tabularnewline
Step& -0.01 &Visit New State & +0.1& $\Delta$ distance to obj & -$\delta$\tabularnewline
 Episode Success & +10 &&&\tabularnewline
\bottomrule

	\end{tabular}
\hfill 
\vspace{-0.3cm}
	\caption{\textbf{Rewards.} In addition to standard RL reward for embodied AI, we include reward components encouraging exploration and efficient manipulation.
	}
	\label{tab:reward}
	\vspace{-0.3cm}
\end{table}

\section{Experiments}
\label{sec:experiment}
We present our experiments comparing \modelShort with baselines (Sec.~\ref{sec:how_well}), a robustness analysis of our model with regards to noise in the agent's motions and sensors (Sec.~\ref{sec:sensor_noise_ablation}) and ablations of \modelShort's design choices (Sec.~\ref{sec:detection_ablation}). Finally we evaluate \modelShort for displacing novel objects, not used during training (Sec.~\ref{sec:zero_shot}).

\noindent \textbf{Baselines:}
We consider the following baselines:
\begin{itemize}[leftmargin=*]
    \item \textbf{No Mask Baseline} -- This model uses the RGBD observation and the query image as input and directly predicts a policy for completing the task. It does not have a segmentation module and does not have access to the agent's location relative to the starting point.
    \item \textbf{Mask Driven Model (MDM)} -- This network uses a very similar architecture as our model with RGBD observation, query images, and the segmentation mask of the target objects as inputs. However, it does not have access to the agent's location relative to the starting point.
    \item \textbf{Location-aware Mask Driven Model (Loc-MDM)} -- This baseline shares the same architecture as MDM. However, it also uses the agent's relative location from the start of the episode. The inputs available to Loc-MDM are the same as our proposed model (\modelShort). The difference is that Loc-MDM uses the agent's location naively, whereas \modelShort\ uses it for target localization.
\end{itemize}

In addition to these baselines, we also train the model from \cite{manipulathor} on our task \taskshort. Note that this model uses the perfect location of the agent and the target objects at each time step, so it is not a fair comparison to our model. However, it is a useful point of reference.
\begin{itemize}[leftmargin=*]
\itemsep-0.2em
   \item \textbf{ArmPointNav}~\cite{manipulathor} -- This method is the same architecture as the one introduced in \cite{manipulathor} with the addition of the query images as input. The network is provided with the perfect location information for the agent and source and destination objects. 
\end{itemize}

\noindent \textbf{Training Details.} We train our models for 20M frames (unless otherwise specified). All the models share the same visual encoder, three convolutional layers to embed the RGBD observation. The maximum episode length is 200 frames, and the episode fails if the agent does not finish the task before the end of the episode. Using \thor~\cite{kolve2017ai2} multi-node training, we render 500 frames per second on 4 machines, each with 4 Tesla T4 GPUs. We use Adam optimizer with a learning rate of 3e-4 and take gradient steps every 128 frames. We consider two objects being in proximity if the centers of objects are less than 20cm apart. We train our conditional segmentation model offline. We use AllenAct~\cite{AllenAct} framework. 
 
\begin{table*}[tp]
\footnotesize
\setlength{\tabcolsep}{2pt}
	\centering
\hfill
	\begin{tabular}{lccccc}
    Model  & Segmentation & Additional Input  & PU &SR& SRwD\tabularnewline
\toprule 
(1) No Mask Baseline& Prediction  & N/A &0.0	&0.0	&0.0\tabularnewline
\midrule
(2) Mask Driven (MDM)& Prediction & N/A & 16.6&	1.62&	0.09 \tabularnewline
\midrule
(3) Loc-MDM& Prediction & Agent's Relative Loc & 20.3&	3.24&	1.08\tabularnewline
\midrule
(4) \modelShort (Ours) & Prediction & Agent's Relative Loc & \textbf{38.7}&	\textbf{11.6}	&\textbf{4.59}\tabularnewline
\midrule
\midrule
(5) Mask Driven (MDM) & GT mask& N/A &50.5&	17.1&	8.74 \tabularnewline
\midrule
(6) Loc-MDM& GT mask & Agent's Relative Loc& 57.3	&21.6&	9.01\tabularnewline
\midrule
(7) ArmPointNav \cite{manipulathor} & N/A &Compass + GPS & 76.6&	58.3&	28.5\tabularnewline
\midrule
(8) \modelShort (Ours) & GT mask& Agent's Relative Loc  & \textbf{81.2} &	\textbf{59.6} &	\textbf{31.0} \tabularnewline
\bottomrule

	\end{tabular}
\hfill 
    \vspace{-0.3cm}
	\caption{\textbf{Quantitative Results.} Rows (1)-(4) present the results with the predicted segmentation masks. Rows (5)-(8) present the results for models when provided with ground truth segmentation. Our model \modelShort outperforms all the baselines across all metrics. The No Mask Baseline is simply unable to train well, in spite of repeated attempts by us to vary hyperparameters and other design choices.}
	\label{tab:exp}
	\vspace{-0.3cm}
\end{table*}

\noindent \textbf{Dataset.}
We use the APND dataset proposed in \cite{manipulathor} to define the tasks' configurations. APND consists of 12 object categories (Apple, Bread, Tomato, Lettuce, Pot, Mug, Potato, Pan, Egg, Spatula, Cup, and SoapBottle), 30 kitchen scenes, 130 agent initial locations per scene, and 400 initial locations per object-pair per room. For each task, we choose two objects, place them in two locations (randomly chosen from APND) and choose one as the source object $O_S$ and the other as the destination $O_D$. We also randomize the initial location of the agent. We split the 30 scenes into 20 for training, 5 for validation, and 5 for testing. We generate 132 object pairs per training scene and 1600 pairs of initial locations of objects per training scene. We select a fixed set of $1100$ tasks evenly distributed across scenes and object categories for validation and test.

We also generate a dataset of images from the training scenes to train our segmentation model. Our dataset consists of more than 23K images, including 116K object masks from 120 scenes (where 80 are used for training, 20 for validation, and 20 for test sets). We ensure that the validation and test scenes used to train \modelShort\ are not used to train the segmentation model. As explained in Section~\ref{sec:task}, the model requires query images as input. We collect these query images by taking a crop containing the object of interest from the images of this dataset. We only use the query images collected from the training scenes of \thor, so the instances depicted in the query images do not overlap with the instances of target objects during inference.

\noindent \textbf{Metrics.}
We use the same evaluation metrics as \cite{manipulathor}:
\begin{itemize}[leftmargin=*, noitemsep]
\item Success rate \textbf{(SR)} -- Fraction of episodes in which the agent successfully picks up the source object ($O_S$) and brings it to the destination object ($O_D$).
\item Success rate without disturbance \textbf{(SRwD)} – Fraction of episodes in which the task is successful, and the agent does not move other objects in the scene. 
\item Pick up success rate \textbf{(PU)} – Fraction of episodes where
the agent successfully picks up the object.
\end{itemize}
\vspace{-1em}

\begin{figure*}[h!]
    \centering
    \includegraphics[width=35pc]{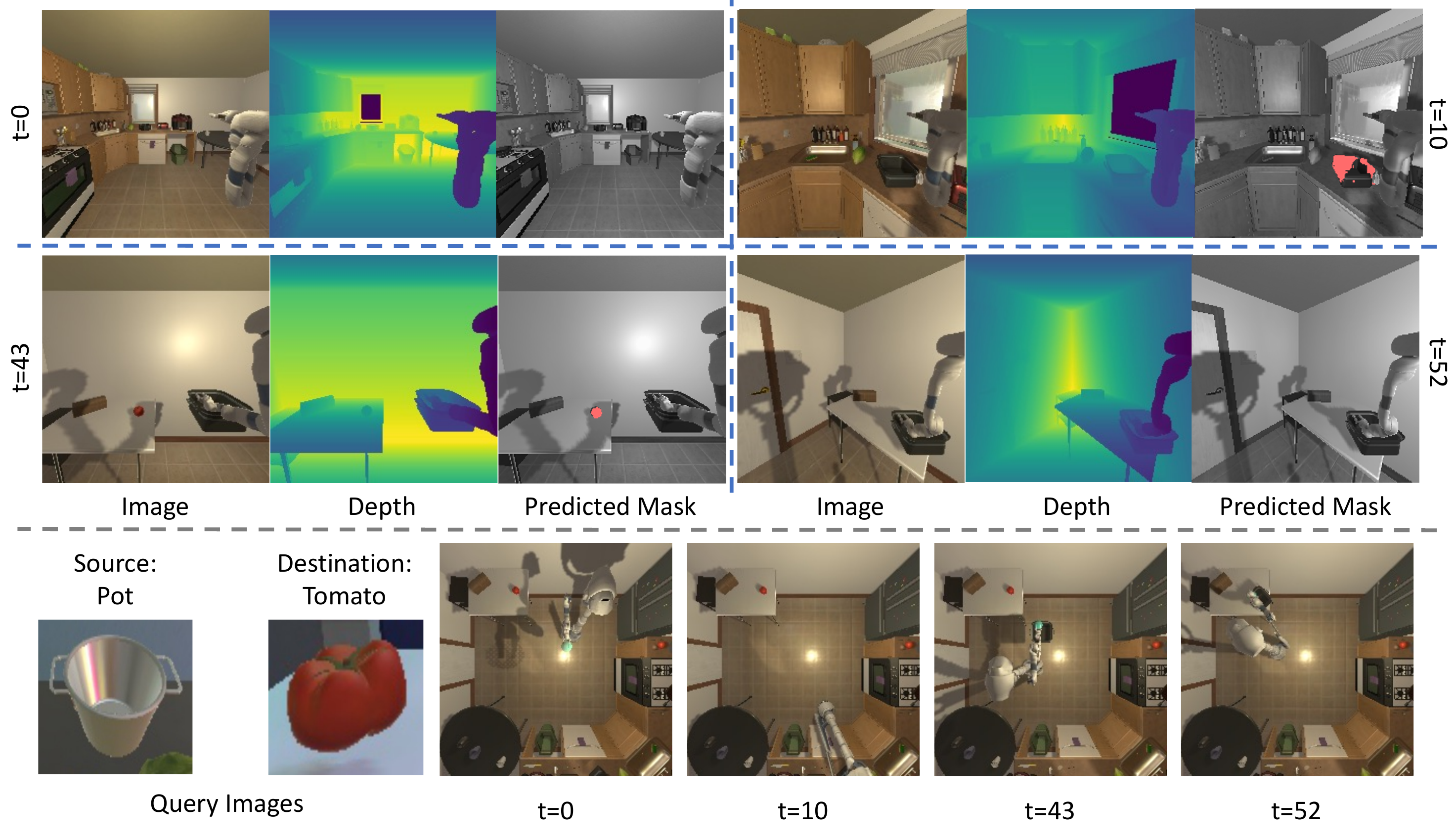}
    \vspace{-0.4cm}
    \caption{\textbf{Qualitative Results}. The figure presents a successful episode of \textit{Bringing a Pot to Tomato}. Despite the errors in Pot segmentation and with only a few pixels of the object being segmented, our agent successfully completes the task. Moreover, towards the end of the episode ($t=52$), the pot in hand occludes the target object (tomato), but the agent is able to remember the target location despite the occlusion. Note that the topdown view (bottom row) is only shown for visualization purposes and is not an input to the network.}
    \vspace{-0.3cm}
    \label{fig:qualitative}
\end{figure*}

\subsection{How well does \modelShort work?}
\label{sec:how_well}
Table~\ref{tab:exp} presents our primary quantitative findings on the test environments. Rows 1-4 provide results for \modelShort\ compared to our baselines. Rows 5-8 evaluate these models when using ground truth segmentation masks in order to assess these models independent of the segmentation inaccuracies.

Our model \modelShort, row 4, outperforms all of the baselines (rows 1-3). Directly learning a policy from input images results in a 0 success rate, showing the advantage of using a separate segmentation and target localization network. This result is despite trying various hyperparameters and designs to get this model off the ground. The improvement of the location-aware methods (rows 3 and 4) over the Mask-driven model (row 2) that does not encode relative location demonstrates that the agents benefit from encoding localization information. Our method \modelShort\ provides a massive 3x improvement over the baseline in row 3 in terms of Success Rate (SR). This result shows that our approach that estimates the object's distance from the agent is quite effective.

Rows 5-8 evaluate models when employing ground truth segmentation masks. \modelShort\ outperforms all baselines significantly, including Loc-MDM with access to the same information as \modelShort. An interesting observation is that \modelShort\ achieves better performance in unseen scenes compared to the ArmPointNav~\cite{manipulathor} baseline (row 7), which receives the relative location of the target object at each time step. \taskName\ requires attending to the visual observations to avoid collisions of the arm with objects in the scene. Therefore, our conjecture for the higher performance is that the ArmPointNav baseline relies heavily on the GPS and location sensors leading to less focus on visual observations. We discuss this further in Appendix~\ref{sec:suppl_apn_vs_us}. Figure~\ref{fig:qualitative} shows our qualitative results. 

One of the main contributions of \modelShort is the ability to maintain an estimation of the target's location regardless of its visibility. To quantify this contribution, we calculated the percentage of the frames in an episode for which the target is not visible during the inference time. Our experiments show that only in $43.9\%$ and $10.5\%$ of the observed frames the source and goal objects are visible in the agent's frame, respectively (mainly due to arm occlusion or object not being in the frame). More importantly, in $70.5\%$ of those frames, the segmentation model fails to segment the object (misdetection). Despite the low percentage of object visibility, \modelShort successfully completes the task by aggregating the temporal information.

\subsection{How robust is \modelShort to noise?}
\label{sec:sensor_noise_ablation}
There are two primary sources of noise that can impact our model, noise in the agent's movements and noise in depth perception. In the following experiments, we evaluate the effect of each one.

\begin{figure}[h!]
    \centering
    \vspace{-0.3cm}
    \includegraphics[width=13pc]{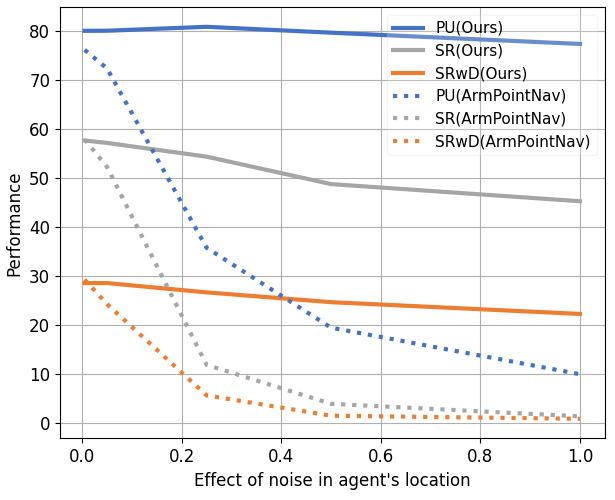}
    \vspace{-0.3cm}
    \caption{\textbf{Robustness to noise in agent's movements}. We use the noise model from \cite{pyrobot2019}. The x axis shows the noise multiplier and the y axis shows the performance on the \taskName\ task. Performance without any noise is shown at $x=0$. $x=1$ corresponds to a noise as large as the agent step size (20cm). }
    \vspace{-0.4cm}
    \label{fig:noise_agent_loc}
\end{figure}

\noindent \textbf{Noise in agent movements.}
We use the motion model presented in \cite{pyrobot2019}, which models the noise in the traveled distance and angle of rotation. The details of the noise model can be found in Appendix~\ref{sec:suppl_noise}. Figure~\ref{fig:noise_agent_loc} shows the robustness of our method to motion noise in comparison with ArmPointNav~\cite{manipulathor} modified to incorporate the same noise model. 
Our approach \modelShort\ is far more robust, whereas ArmPointNav degrades significantly. We believe our method better leverages visual information to recover from noisy estimates.

\begin{table}[tp]
\footnotesize
\setlength{\tabcolsep}{4pt}
	\centering
\hfill
	\begin{tabular}{rcccc}
    Model  & PU &SR& SRwD\tabularnewline
\toprule
MDM @100M & 20.5 (-40.0\%)&	3.5 (-71.5\%)&	2.07 (-72.6\%)\tabularnewline
\midrule
Loc-MDM @114M & 25.3 (-27.3\%)	&4.05 (-64.8\%)	&1.89 (-75.1\%)\tabularnewline
\midrule
\modelShort(Ours) @20M &\textbf{36.3 (-6.2\%)}	&\textbf{6.13 (-47.7\%)}	&\textbf{2.52 (-45.0\%)} \tabularnewline
\bottomrule
	\end{tabular}
\hfill 
\vspace{-0.3cm}
	\caption{\textbf{Robustness to noise in depth.} Our model achieves the best performance on all metrics and the lowest relative performance drop compared to the baselines (shown in parentheses).}
	\label{tab:depth_noise}
	\vspace{-0.3cm}
\end{table}

\noindent \textbf{Noise in depth perception.}
We use the Redwood depth noise model presented in \cite{choi2015robust}. Table~\ref{tab:depth_noise} shows the performance of models in the presence of this noise. Note that models are trained with no noise, but inference takes place in the presence of noise. In this experiment, we are not only interested in evaluating models' absolute performance in the presence of noise but also in the relative degradation of each model. With this in mind, to have a fair comparison, we train all models until they reach the same success rate on the training set as \modelShort. We then evaluate them with and without depth noise and calculate the relative performance drop in all metrics. As seen in Table~\ref{tab:depth_noise}, \modelShort\ is more robust to noise in depth compared to our baselines (in absolute terms) and also shows a smaller relative performance degradation. We hypothesize that the aggregated information throughout the episode helps our model recover from the locally observed noise in depth. 

\begin{table}[tp]
\footnotesize
\setlength{\tabcolsep}{4pt}
	\centering
\hfill
	\begin{tabular}{lccc}
    Model & PU & SR & SRwD \tabularnewline
\toprule 
 Our Policy with MaskRCNN~\cite{He2017MaskR} detection & 31.3&	8.7&	3.87 \tabularnewline
\midrule
Our Policy with Conditional Segmentation & \textbf{38.7}&	\textbf{11.6}&	\textbf{4.59} \tabularnewline
\bottomrule

	\end{tabular}
\hfill 
\vspace{-0.2cm}
	\caption{\textbf{Instance segmentation ablation.} We ablate the performance of our model using different instance segmentation methods. To do so we train a MaskRCNN~\cite{He2017MaskR} model on our data and show that our conditional segmentation helps our policy achieve better performance on the \taskshort\ task. }
	\label{tab:detection}
	\vspace{-0.3cm}
\end{table}

\subsection{Why Conditional Segmentation?}
\label{sec:detection_ablation}
Simultaneously learning to segment objects and learning to plan is challenging. Hence, we isolate the segmentation branch of our network. Most approaches to ObjectNav~\cite{batra2020objectnav} (navigating towards a specified object) do not use external object detectors and learn to detect and plan jointly, but that is not effective for our task. As we showed in Table~\ref{tab:exp}, the baseline that does not use a separate segmentation network (row 1) generalizes poorly and achieves 0 success rate.

There are two advantages to our conditional segmentation method compared to the standard state-of-the-art detection/segmentation models such as MaskRCNN~\cite{He2017MaskR}. First, our task is more straightforward since we are interested in simply segmenting an instance of the specified object category. Thus, we can afford to train a much smaller network. Second, as shown in Table~\ref{tab:detection}, the model that uses conditional segmentation obtains better performance in all metrics. For this experiment, we use the MaskRCNN~\cite{He2017MaskR} model pre-trained on the LVIS dataset~\cite{gupta2019lvis} and finetuned on \thor\ images. We obtain the highest confidence prediction for the target class from MaskRCNN and use that as the input to our policy. This is effective because there is typically just one object instance of the specified category in view.

\begin{table}[h!]
\footnotesize
\setlength{\tabcolsep}{4pt}
	\centering
\hfill
	\begin{tabular}{lcccc}
    Model & Object Set & PU & SR & SRwD \tabularnewline
\toprule 
Loc-MDM & NovelObj & 9.52&	0.9&	0 \tabularnewline
\midrule
\modelShort & NovelObj &  \textbf{26.2}&	\textbf{5.24}&	\textbf{3.33}\tabularnewline
\midrule
\midrule
Loc-MDM & SeenObj& 29.3&	5.67&	2.33\tabularnewline
\midrule
 \modelShort & SeenObj & 49.3&	18.7&	10.7 \tabularnewline
\bottomrule

	\end{tabular}
\hfill 
\vspace{-0.3cm}
	\caption{\textbf{Zero-shot Manipulation Results.} }
	\label{tab:zero_shot}
\vspace{-0.5cm}
\end{table}

\subsection{Can \modelShort do zero-shot manipulation?}
\label{sec:zero_shot}
So far, we have evaluated our model on generalization to unseen scenes. We evaluate whether the model can manipulate novel objects not used for training. This is a challenging task since the variation in object size and shape can result in different types of collisions and occlusions. For this evaluation, we train our model on six object categories (Apple, Bread, Tomato, Lettuce, Pot, Mug) and evaluate on the held-out classes (Potato, Pan, Egg, Spatula, Cup, SoapBottle). Note that our conditional segmentation model that provides the segmentation input to the policy has been trained on all categories. However, our policy has not been exposed to the test objects. Table~\ref{tab:zero_shot} shows the results. Our model generalizes well to the novel categories, which is challenging as the novel objects' shapes and sizes can result in unseen patterns of collision.

\section{Conclusion}
We propose a method for visual object manipulation, where the goal is to displace an object between two locations in a scene. Our proposed approach learns to estimate target object location via aggregating noisy observations caused by missed detection, view occlusion by the arm, and noisy depth. We show that our approach provides a 3x improvement in success rate over a baseline without this auxiliary information, and it is more robust against noise in depth and agent movements. 

{\textbf{Acknowledgement.} We thank Luca Weihs and Winson Han for their valuable input on the project, Jordi Salvador for his impactful contribution on multinode training and Klemen Kotar for his help with training the off-the-shelf detection model.}
{\small
\bibliographystyle{ieee_fullname}
\bibliography{egbib}
}

\clearpage
\appendix

\section{Training Details for Conditional Segmentation Model -- Sec \ref{sec:condseg}}
We train our conditional segmentation models for 100 epochs. We use Adam optimizer with a learning rate of $1e-4$. We initialize the ResNet backbones with ImageNet pre-trained weights for faster training. Input images and query images are of size 224x224. We use random crop and flipping the image for data augmentation during training. We do not use color jittering as we found it to negatively impact the performance of our model on the validation set. 

\section{Reward Ablation -- Sec \ref{sec:policy}}
\label{sec:reward_ablation}
Section \ref{sec:policy} introduced three additional reward elements that are not used in the standard ArmPointNav setup. We ablate the performance gain that each component brings to our model. This analysis can be useful not only for evaluating the performance of our method but also for use in future works. Table~\ref{tab:reward_ablation} includes the results in absence of each of these components. Arm control reward, which encourages the agent to move its arm close to the target objects, is shown to be the most important element of the reward and the agent does poorly without this reward. Getting training off the ground with sparse RL reward for object manipulation is extremely difficult, and by motivating the agent to bring its arm closer to the target object, the training becomes faster and more efficient.

\begin{table}[h!]
\footnotesize
\setlength{\tabcolsep}{4pt}
	\centering
\hfill
	\begin{tabular}{lccccc}
    Exploration & Arm Control &Object Visibility& PU & SR & SRwD\tabularnewline
\toprule 
 \checkmark & - & \checkmark &  0.9	&0&	0\tabularnewline
\midrule
 - & \checkmark & \checkmark & 47.8&	15.4&	5.86\tabularnewline
 \midrule
 \checkmark & \checkmark & - & 79.4	&53.6&	30.1 \tabularnewline
\midrule
 \checkmark & \checkmark & \checkmark & \textbf{81.2}&	\textbf{59.6}&	\textbf{31.0} \tabularnewline
\bottomrule

	\end{tabular}
\hfill 
\vspace{0.1cm}
	\caption{\textbf{Reward ablations.} We investigate the performance of the model in absence of each reward component. Exploration, arm control, and object visibility refer to the rewards for visiting a new state, $\delta$ distance to object and observing the object, respectively. }
	\label{tab:reward_ablation}
\end{table}

\section{Details of the Noise Models -- Sec \ref{sec:sensor_noise_ablation}}
\label{sec:suppl_noise}
This section provides additional details and visualizations on the noise models used in Section \ref{sec:sensor_noise_ablation}. 
\subsection{Agent Motion Noise Model}

\begin{figure}[h!]
    \centering
    \includegraphics[width=15pc]{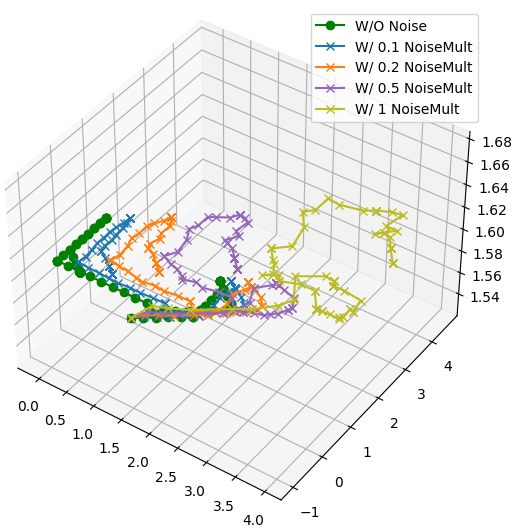}
    \caption{\textbf{Noise in Trajectory.} The green path shows the trajectory traversed by the agent, and the other paths show the estimated trajectory of the agent for a variety of noise multipliers. Note that even a small noise multiplier $x=0.2$ can result in a big divergence in the trajectories.}
    \label{fig:trajectory_noise}
\end{figure}

Murali et al.~\cite{pyrobot2019}, introduced a noise model for agent's movements and rotation, capturing the noise in motion for a real robot, the Locobot. We use their noise model to ablate the impact of the noise in the agent's movements on our model's performance. To better understand how the noise multiplier (the x-axis in Figure 5 in the main submission) impacts the predicted trajectory, we illustrate a sample trajectory. Figure~\ref{fig:trajectory_noise} shows agent's groundtruth and predicted trajectory for different noise multipliers $x=0.1-1$. Note that the paths start from the same initial location, and the predicted paths diverge from the actual trajectory traversed by the agent as the episode progresses and the errors accumulate.

\subsection{Depth Noise Model}

Redwood distortion noise model, introduced in \cite{choi2015robust}, is designed to model the noise of actual depth sensors (Kinect cameras) with a fixed resolution. This noise model can act as a proxy to estimate the impact of using a real-world noisy depth camera. 

\subsection{How Does Segmentation Affect the Final Performance? -- Sec \ref{sec:detection_ablation}}

We compare the performance of our approach using different segmentation models in Section \ref{sec:detection_ablation}. In this section, we ablate how the performance of the segmentation model affects the final performance of our approach on the task of Object Displacement.

\begin{figure}[h!]
    \centering
    \includegraphics[width=15pc]{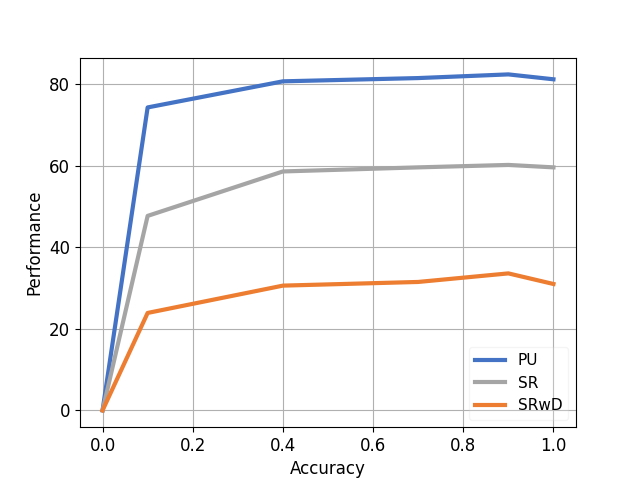}
    \caption{\textbf{Impact of Partial Masks.}}
    \label{fig:abl2}
\end{figure}

First, we evaluate how the performance on the final task changes if the target objects are detected correctly, but the mask does not perfectly segment the observed object. In other words, if we use IoU as a metric for the accuracy of the segmentation model, how the accuracy of the segmentation network affects the final performance. For this experiment, at each timestep that the object is visible, we randomly choose $x\%$ of the segmentation mask of the target object to preserve and remove the rest. 
Figure~\ref{fig:abl2} plots the change in the final performance as we increase $x$. Note that $x=0$ presents the ablation where no mask is provided, and $x=1$ is equivalent to providing the groundtruth segmentation mask. This experiment shows that our network can achieve strong results even if only $10\%$ of the target object is segmented. 

\begin{figure}[h!]
    \centering
    \includegraphics[width=15pc]{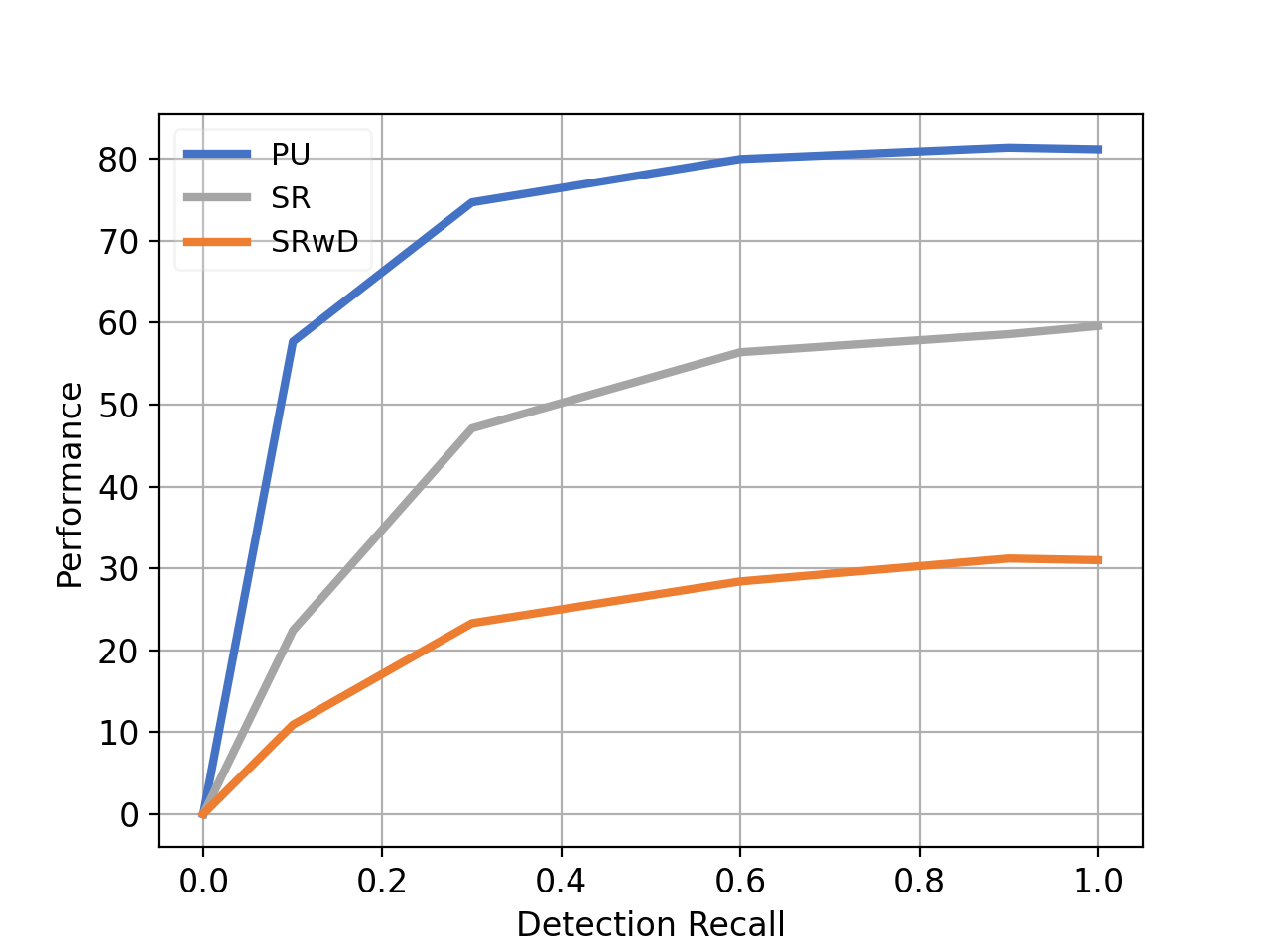}
    \caption{\textbf{Impact of Missing Masks.}}
    \label{fig:abl1}
\end{figure}

We also show that if the segmentation mask of the object is only retrieved $30\%$ of the times, across all the timesteps that the target object is visible, our network can achieve approximately similar performance as the one using the ground-truth segmentation mask (Figure~\ref{fig:abl1}). For every $x$ on the plot, at each timestep, we remove the segmentation mask of the target object with the probability $1-x$ and calculate the final performance of the model. Similar to the previous plot, $x=0$ shows the performance when no mask is provided, and $x=1$ is equivalent to the groundtruth segmentation mask. 

\begin{figure}[h!]
    \centering
    \includegraphics[width=15pc]{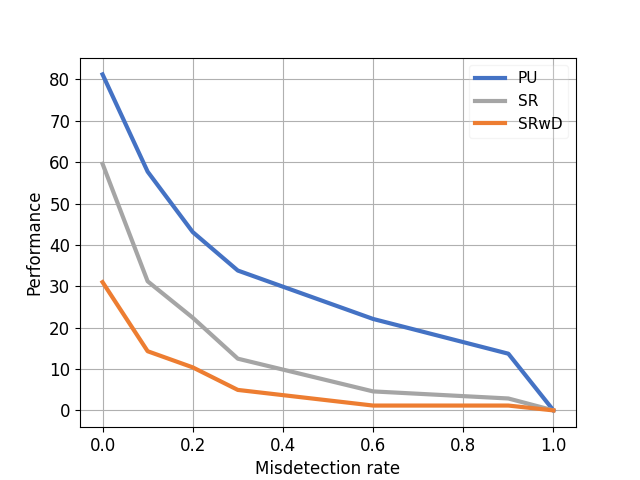}
    \caption{\textbf{Impact of Category Confusion.}}
    \label{fig:abl3}
\end{figure}

One of the other main issues that the segmentation models have is detecting wrong objects. For instance, as shown in the supplementary video, the segmentation model might segment a pan in the scene while the query object asks for a pot. Figure~\ref{fig:abl3} shows the final performance of the model for different rates of mis-detections. At each timestep, with the probability of $x\%$, instead of the segmentation mask of the target object, we randomly select the segmentation mask of another object in the scene as the input mask to the model. For instance, metrics at $x=0.2$ show the model's performance using a segmentation network that detects a wrong object with the probability of $20\%$.

The conclusion is that even segmentation models with high precision and low recall are beneficial for the Object Displacement task.

\section{Our method (\modelShort) outperforming ArmPointNav (APN). -- Sec \ref{sec:how_well}}
\label{sec:suppl_apn_vs_us}

\begin{figure}[h!]
    \centering
    \includegraphics[width=20pc]{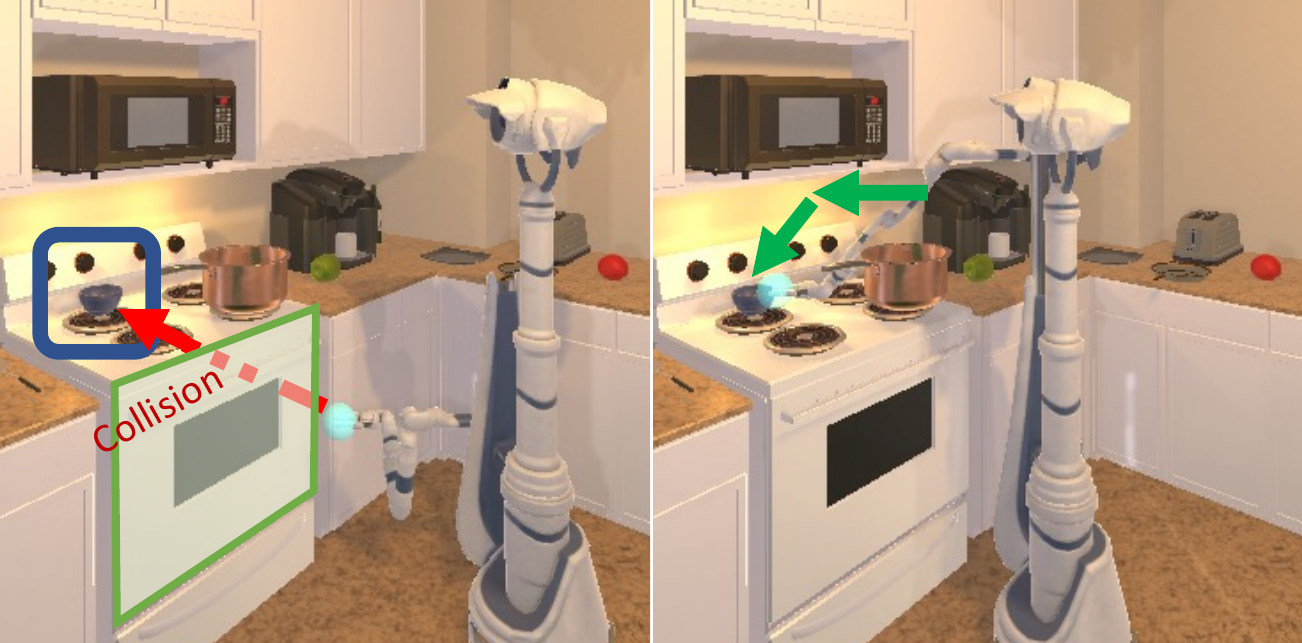}
    \caption{\footnotesize \textbf{\modelShort vs APN paths towards the target.} The red path illustrates the greedy solution an agent with access to GT direction sensors might take to reach the bowl (Failing due to collision with the oven).
    }
    \label{fig:combined}
\end{figure}

APN has access to the exact direction towards the target, so it tends to choose the direct path towards the goal, which is not necessarily a plausible solution. For example, as shown in Fig.~\ref{fig:combined}, the red path represents a shorter path, but the arm collides with other objects if it follows that path. 
Moreover, we did a further quantitative analysis to investigate \modelShort superiority to APN (Tab~\ref{tab:apn_vs_ours}). EpLen PU and EpLen Success, respectively, represent episode length for the pickup stage and the full task. Our analysis shows that these metrics are lower for APN than \modelShort. However, the SRwD of APN (success without collision) is lower. This observation shows that APN is more efficient in the number of steps (as it uses the exact direction) but does not handle collisions well. 

\begin{table}[h!]
\footnotesize
\setlength{\tabcolsep}{4pt}
	\centering
\hfill
	\begin{tabular}{lccc}
    Model  & EpLen PU& EpLen Success & SRwD \tabularnewline
\toprule 
ArmPointNav &46.3 & 78.8& 28.5\tabularnewline
\modelShort w/ GT mask & 49.3 & 87.1 & 31.0\tabularnewline
\bottomrule
	\end{tabular}
\hfill 
	\vspace{0.1cm}
	\caption{\footnotesize \textbf{\modelShort vs APN.}}
	\label{tab:apn_vs_ours}
\end{table}

\section{Limitations}
Here, we discuss two main limitations of the work. First, we consider separate modules for segmentation and depth perception. Hence, their errors have a cascading effect. Another design choice we made was to abstract away grasping. While this allowed us to focus on other challenging aspects of the problem, moving beyond simulation will likely require methods to account for real-world graspers. These are some of our study's primary limitations, which serve as avenues for future work.

\end{document}